\documentclass{spie}

\usepackage{cite}

\usepackage{times} 
\usepackage{indentfirst} 
\usepackage{amsmath}
\usepackage[hidelinks]{hyperref} 
\hypersetup{bookmarks=true, colorlinks=true, linkcolor=blue, citecolor=blue, urlcolor=blue, pdfborder={0 0 0.5}} 

\usepackage[colorinlistoftodos,color=pink]{todonotes} 
\usepackage{xkeyval}
\presetkeys{todonotes}{inline}{}

\usepackage{lipsum} 
\usepackage{subcaption} 
\usepackage{caption}

\usepackage{ragged2e}
\justifying

\title{Infrared Domain Adaptation with Zero-Shot Quantization}

\author{Burak Sevsay, Erdem Akagündüz
\skiplinehalf
\normalsize 
Graduate School of Informatics, Middle East Technical University, Ankara, Türkiye\\
{\tt\small \{burak.sevsay, akaerdem\}@metu.edu.tr}}

\usepackage{caption}
\begin{document}

\maketitle

\begin{abstract}
Quantization is one of the most popular techniques for reducing computation time and shrinking model size. However, ensuring the accuracy of quantized models typically involves calibration using training data, which may be inaccessible due to privacy concerns. In such cases, zero-shot quantization, a technique that relies on pretrained models and statistical information without the need for specific training data, becomes valuable. Exploring zero-shot quantization in the infrared domain is important due to the prevalence of infrared imaging in sensitive fields like medical and security applications. In this work, we demonstrate how to apply zero-shot quantization to an object detection model retrained with thermal imagery. We use batch normalization statistics of the model to distill data for calibration. RGB image-trained models and thermal image-trained models are compared in the context of zero-shot quantization. Our investigation focuses on the contributions of mean and standard deviation statistics to zero-shot quantization performance. Additionally, we compare zero-shot quantization with post-training quantization on a thermal dataset. We demonstrated that zero-shot quantization successfully generates data that represents the training dataset for the quantization of object detection models. Our results indicate that our zero-shot quantization framework is effective in the absence of training data and is well-suited for the infrared domain.

\end{abstract}  
\section{Introduction}
\label{sec:intro}

Quantization is an emerging technique in the deployment of neural networks because of the growing need for faster computation and a smaller memory footprint. However, quantization to 8-bit and below precision leads to significant quantization error, causing a notable drop in accuracy. In practice, Post Training Quantization (PTQ) and Quantization Aware Training (QAT) are the two most common techniques to apply quantization while minimizing accuracy loss. Unlike PTQ, QAT includes the retraining process of the quantized model to recover the accuracy drop. However, in many cases, training data may not be available due to privacy concerns, making QAT infeasible. In such cases, PTQ stands out as the most appropriate option for quantization because PTQ can be applied without the training data.


\begin{figure}[t]
  \centering
    \captionsetup{justification=centering, font={normalsize}, width=\textwidth}
  \begin{subfigure}[b]{0.25\textwidth}
    \includegraphics[width=\linewidth]{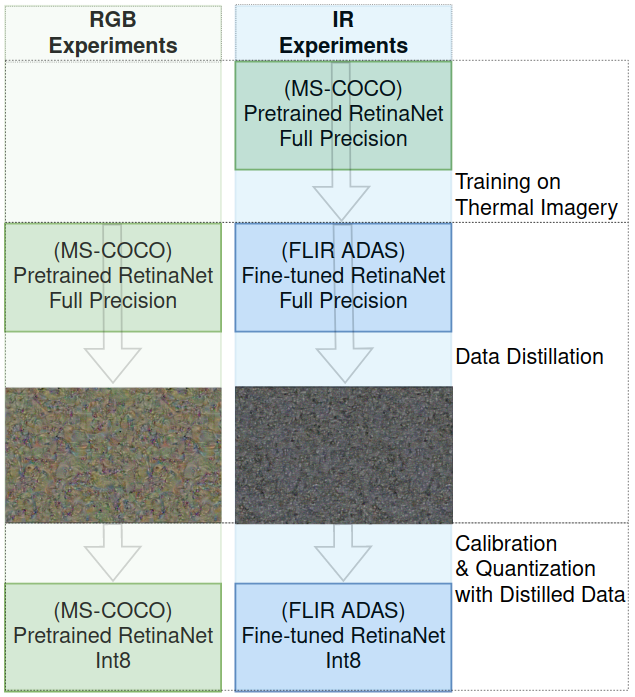}
    \caption{The proposed zero-shot quantization framework}
    \label{fig:system_design}
  \end{subfigure}
  \hspace{0.9cm} 
  \begin{subfigure}[b]{0.5\textwidth}
    \includegraphics[width=\linewidth]{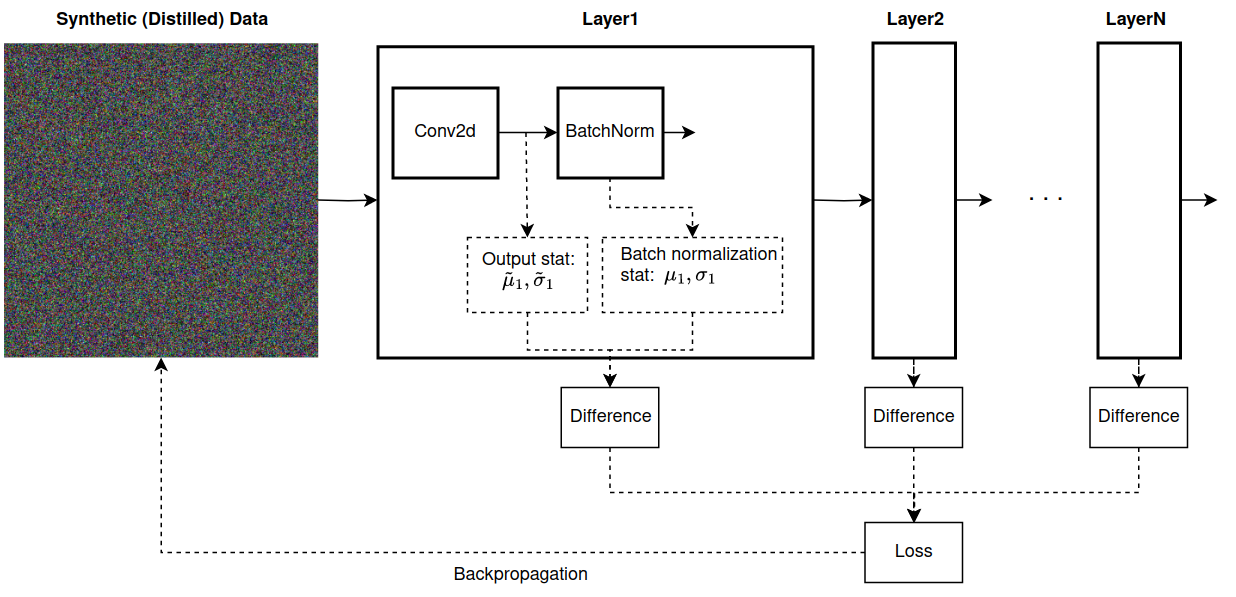}
    \caption{Data distillation process for zero-shot quantization}
    \label{fig:distillation-process}
  \end{subfigure}
  \caption{Overview of the zero-shot quantization process}
  \label{fig:combined}
\end{figure}

There are several determinations that need to be made for the PTQ process. The initial consideration is choosing between uniform \cite{q_Lin2015FixedPQ, qat_Jacob2017QuantizationAT, ZeroQ} and non-uniform quantization \cite{nonuniform_jung2018learning}. Non-uniform quantization may yield higher accuracy due to its higher representational capacity, but uniform quantization proves to be more resource-efficient in terms of memory footprint and computation speed. Another important aspect of quantization is the selection of the clipping range of the model parameters. The process is called calibration. The straightforward method of calibration is finding the minimum and maximum values within the selected parameter group. Also, techniques like percentile and entropy can be used to determine the clipping range. The clipping range can be determined layer-wise, group-wise, or channel-wise. The usage of smaller groups as granularity can enhance accuracy but may extend the required time of the calibration.

After outlining the specifics of post-training quantization (PTQ), one method to implement it without any training data is through dynamic quantization \cite{dynamic_Liu_2022_CVPR}. In this method, the clipping range is determined dynamically in the inference of each input data. It results in higher accuracy than static quantization, but computing the clipping range for each input is not always a feasible solution for real-time requirements. Dynamic clipping range computation slows down the inference process, which limits the main benefit of quantization. On the other hand, the clipping range is determined only once in static quantization. It is a better option to reduce inference time and memory footprint. However, training data is required to capture the most suitable clipping range in static quantization.

Zero-shot quantization \cite{ZeroQ,adv_Liu2021ZeroshotAQ,KD_Nayak2019ZeroShotKD,bns_Xu2020GenerativeLD,bns_Zhang2021DiversifyingSG} addresses this problem and makes static quantization available without any training data. Zero-shot quantization aims to generate synthetic data to determine the clipping range. One of the most common approaches is the usage of batch normalization statistics to distill synthetic data. Losses in this technique are based on the difference between batch normalization statistics and previous layer output statistics. The loss is propagated to the input image to decrease the gap between statistics. The resulting image can be used in calibration instead of training data.

Infrared domain applications require zero-shot quantization due to the private nature of the datasets primarily used in medical, security, or autonomous driving applications. In our work, we investigated the performance of zero-shot quantization for the infrared domain. Initially, we applied zero-shot quantization to an MS-COCO pre-trained RetinaNet as a baseline experiment. Subsequently, we fine-tuned the pre-trained RetinaNet using the FLIR ADAS dataset for the object detection task and showed that zero-shot quantization is applicable to thermal imagery. We presented distilled data, accuracy results, and model size comparison between the COCO pre-trained RetinaNet and the FLIR ADAS fine-tuned RetinaNet to explore the impact of thermal imagery on zero-shot quantization. Our experiment scheme within the proposed zero-shot quantization framework is illustrated in Figure \ref{fig:system_design}. 

Moreover, in our experiments, we compared the use of only the mean statistic with the use of both the mean and standard deviation statistics for the loss function. We further demonstrated that distilled data successfully represents the training data for quantization performance. Thus, zero-shot quantization emerges as an important alternative to using training data for quantization.

\section{Related Work}
\label{sec:relatedWork}

The use of real-time computation and edge devices increases the need for model compression. Increasing model size and complexity also increases the compression requirements. Quantization \cite{ q_Lin2015FixedPQ, q_Liu2021PostTrainingQF, q_Wu2015QuantizedCN, bnn_Rastegari2016XNORNetIC, tnn_Alemdar2016TernaryNN, qat_Jacob2017QuantizationAT} is one of the most popular choices for model compression but there are also other model compression techniques such as model pruning \cite{Prune_Li2016PruningFF}, knowledge distillation \cite{KD_Hinton2015DistillingTK} and efficient neural model architecture design \cite{ENAS_Ding2021BNASEN}. Additionally, these methods can be combined \cite{mix_Aghli2021CombiningWP,mix_Kim2021PQKMC} to increase model compression. 

Model quantization reduces the bit precision of a model's weights and activations. Quantization limits can be pushed with binary neural networks \cite{bnn_Rastegari2016XNORNetIC} and ternary neural networks \cite{tnn_Alemdar2016TernaryNN} which may be beneficial for FPGA deployment of neural networks. The most challenging part of quantization is minimizing accuracy drop because of quantization error. Therefore, quantization requires a tuning algorithm to reduce accuracy drop. Retraining of quantized models \cite{qat_Jacob2017QuantizationAT, qat_Tailor2020DegreeQuantQT} can be successful in minimizing accuracy drop but the process requires a training dataset to fine-tune the quantized model. Clipping outliers \cite{ocf_Zhao2019ImprovingNN}, which makes the quantization process more effective is another method that requires training data. 

Quantization methods that do not require any training/validation data are called zero-shot quantization methods \cite{survey_Gholami2021ASO}. Weight equalization and bias correction \cite{webc_Nagel2019DataFreeQT} are data-free quantization techniques to apply. Additionally, there are techniques to generate synthetic data for quantization without accessing any training data. Generating adversarial samples \cite{adv_Choi2021QimeraDQ, adv_Liu2021ZeroshotAQ, adv_Zhu2021AutoReConNA} is one way to generate synthetic data for quantization. Furthermore, batch normalization statistics can be utilized for synthetic data generation. ZeroQ \cite{ZeroQ} proposed a data distillation method that is based on batch normalization statistics. Also, there are other zero-shot quantization methods, mostly limited to classification tasks, based on batch-normalization statistics \cite{bns_Xu2020GenerativeLD, bns_Zhang2021DiversifyingSG, hardsamples, intraq}. This paper follows a data distillation procedure similar to that in \cite{ZeroQ}.   

Object detection in the infrared domain is an active research area \cite{thermal_general_10208595,thermal_general_7298706,thermal_fuse_Devaguptapu_2019, thermal_general_9133581}. There are methods that fuse visual and thermal features from data \cite{thermal_fuse_Devaguptapu_2019, thermal_fuse_Knig2017FullyCR, thermal_fuse_Wagner2016MultispectralPD}. Additionally, training and fine-tuning detection networks are quite popular \cite{thermal_finetune}. In application fields like automotive, medical, and security, real-time computation requirements for thermal imagery are crucial. Some works have focused on the quantization of models trained or fine-tuned with thermal images \cite{thermal_quant_1, thermal_quant_9576741, thermal_quant_s23218723}. However, all of these studies require training data. To our knowledge, the performance of zero-shot quantization in the infrared domain has not been investigated. Our main contribution is exploring the performance of zero-shot quantization on models fine-tuned with thermal images.

\section{Methodology}
\label{sec:methodology} 

\subsection{Basic Quantization}

We use uniform asymmetric quantization (aka the uniform affine quantization) in our implementation. Asymmetric quantization generally captures a tighter clipping range since the distributions of activation and weight values are not generally centered around zero. The first part of asymmetric quantization is determining the clipping range. There are many methods \cite{calib_choukroun2019lowbit, calib_mckinstry2019discovering} to determine the clipping range, such as min/max, percentile and entropy methods. We use the percentile method with a 99.99\% threshold for the clipping range because it performed better than the others in our experiments.

To capture the appropriate clipping range for quantization, the most commonly used method is forwarding a batch of training data through the network. In zero-shot quantization, which is explained in section \ref{sec:zero-shot-quant}, synthetic data is used instead of training data to determine the clipping range. Additionally, we use training data in a part of our experiments for comparison purposes. The usage of training data for calibration is a typical example of post-training quantization.

Starting from the real-valued tensor r, we used the clipping range of [a,c] where \(a = min(r), c = max(r)\) . Then, we apply quantization to discretize the value distribution into even intervals. There are three parameters to perform quantization, \textit{the scale factor s, the zero-point z, and the bit-width k}. For k-bit representation, quantization maps the parameters to \([-2^{k-1}, 2^{k-1} - 1]\). Since we use uniform quantization, the clipping range [a,c] is divided into \(2^k -1\) uniform intervals. The interval length is represented by scaling factor which can be calculated as \( s = \frac{{c - a}}{{2^k - 1}} \). Since the method is asymmetric, we need a zero-point to map real value zero to integer value which is \(z = -\text{round}\left(\frac{a}{s}\right) - 2^{k-1}\). With these parameters, quantization operation \(Q(.)\) is formulated in \ref{eq:quant-eq}. Rounding operation \(round(.)\) rounds the input value to the nearest integer. Also, the resulting value is clamped between \([-2^{k-1}, 2^{k-1} - 1]\).

\begin{equation}
    Q(x) = \text{clamp}\left(\text{round}\left(\frac{x}{s}\right) + z\right)
    \label{eq:quant-eq}
\end{equation}

\subsection{Zero-Shot Quantization}
\label{sec:zero-shot-quant}
In the absence of training data, zero-shot quantization is an emerging data-free quantization technique that focuses on generating synthetic image for the calibration process of quantization. Our data distillation methodology is based on ZeroQ \cite{ZeroQ}, which uses batch normalization statistics to distill data. We visualized the data distillation process in Figure \ref{fig:distillation-process}. Data distillation is applied to the full precision pretrained model, which is RetinaNet \cite{retinanet}, in our experiments. 

Data distillation is applied to both the RGB image-trained model and the thermal image-trained models. Since the only change in the architecture is the output number of the classification head, the same distillation process is applied to both models, and a three-channel image is distilled. This leads to a difference in the comparison of zero-shot quantization and PTQ. For the thermal image-trained model, the training dataset contains 8-bit single-channel images, but the distilled data is 24-bit three-channel images. For a fair comparison, training data is fed as three channels with duplication of the data in PTQ.

Each layer that contains a batch normalization layer can be used for loss calculation in data distillation. The difference between statistics (mean and standard deviation) stored in batch normalization and the statistics of the output of preceding layer is used for loss calculation. The loss is backpropagated to the input image, which is initialized randomly from a Gaussian distribution with zero mean and unit variance. The main objective is to optimize the input data to match statistical distributions. The optimization problem is given in Formula \ref{eq:zeroq-distill} where $x^d$ is the input data to be distilled, $\tilde{\mu}^d_i / \tilde{\sigma}^d_i$ are mean/standard deviation of distribution of distilled data at $i-th$ batch normalization layer, and $\ \mu_i / \sigma_i$ are the mean/standard deviation parameters stored in the $i-th$ batch normalization layer.  

\begin{equation}
\min_{\mathbf{x}^d} \frac{1}{L} \sum_{i=0}^{L} \left( \lVert \tilde{\mu}^d_i - \mu_i \rVert^2_2 + \lVert \tilde{\sigma}^d_i - \sigma_i \rVert^2_2 \right)
\label{eq:zeroq-distill}
\end{equation}

\section{Experiments}
\label{sec:experiments} 

\subsection{Setup}
A system equipped with an NVIDIA RTX-3090Ti GPU is utilized for the experiments. The entire training phase of the experiments consumed approximately 300 GPU hours.
\subsubsection{Dataset}
FLIR ADAS \cite{fliradasdataset}, as a thermal dataset, and MS-COCO \cite{lin2015microsoft}, as an RGB dataset, are utilized in the experiments. MS-COCO is used for only evaluation.
The FLIR ADAS dataset v1.3 is utilized for both training and evaluation purposes. It comprises both thermal and RGB images, yet we exclusively employed thermal images in our experiments. The thermal images were captured by Teledyne FLIR Tau 2, featuring a resolution of 640x512. The dataset encompasses 10,228 images (8,862 training images, 1,366 validation images), including instances of persons, cars, and bicycles categories. Due to an insufficient number of dog instances in both the training and validation sets, we have removed the dog annotations from the dataset. 

\subsubsection{Model}
We used the PyTorch \cite{paszke2019pytorch} implementation of RetinaNet \cite{retinanet} (v2) for all experiments. It has ResNet-50-FPN backbone. Additionally, we employed PyTorch weights for the MS-COCO pretrained RetinaNet model. In the experiments involving the FLIR ADAS dataset, the RetinaNet classification head was modified for three classes output: people, cars, and bicycles. Since thermal images consist of 8-bit one-channel data, we fed them into the model by duplicating the data across three input channels.

\subsection{Implementation Details}
\subsubsection{Training}
We fine-tuned RetinaNet using the FLIR ADAS dataset. All layers are trainable throughout the fine-tuning process. Parameters for the fine-tuning training are selected to be consistent with the original RetinaNet paper. We use stochastic gradient descent (SGD) with an initial learning rate of 0.01, weight decay of 0.0001, and momentum of 0.9. The learning rate is divided by 5 after every 10 epochs. A batch size of 8 images is utilized, and the training lasts for 40 epochs. 
\subsubsection{Zero-Shot Quantization}
All weights and activations of RetinaNet are quantized to 8 bits. We utilized TensorRT's \cite{Nvidia} PyTorch-Quantization toolkit for the quantization process, which is conducted using distilled data. The hyper-parameters for data distillation, initially set for the RGB experiments, were also used in the infrared experiments and were consistent with the ZeroQ \cite{ZeroQ} implementation. We used an initial learning rate of 0.1 with the Adam optimizer \cite{adam_optim}. For experiments that utilized mean and standard deviation loss in distillation, we divided the learning rate by 5 at 20 and again at 75 iterations.
\subsubsection{Post-Training Quantization}
We investigated the performance of the FLIR ADAS dataset for post-training quantization. We divided the training dataset into batches of size 8 and used these batches for calibration during post-training quantization. We then compared the results with those from zero-shot quantization. 
\subsubsection{Evaluation}
For evaluation metrics, the standard mAP 0.5:0.05:0.95 metric on the COCO dataset is utilized.

\subsection{Results}
\begin{table*}[htbp]
    \centering
      \captionsetup{justification=centering, font={normalsize}, width=\textwidth}
    \caption{Accuracy comparison of full precision (32-bit weights, 32-bit activations) and quantized (8-bit weights, 8-bit activations) RetinaNet with pretrained weights and weights after fine-tuning. Zero-shot quantization is applied.
}
    \label{tab:acc-size-results}
    \begin{tabular}{|c|c|c|c|c|}
        \hline
         & Evaluation Dataset & Precision-bit 

 & mAP & Model Size (MB) \\
        \hline
        Pretrained RetinaNet (FP) & Microsoft COCO & W32-A32 & 41.6 & 145.7 \\
        \hline
        Pretrained RetinaNet (Quantized) & Microsoft COCO & W8-A8 & 40.1 & 36.4 \\
        \hline
        Fine-tuned RetinaNet (FP) & FLIR ADAS & W32-A32 & 32.3 & 138.8 \\
        \hline
        Fine-tuned RetinaNet (Quantized) & FLIR ADAS & W8-A8 & 31.2 & 34.7 \\
        \hline
    \end{tabular}

\end{table*}


Our experiments resulted in three main comparisons, detailed in the following sections. The first comparison is between the full-precision model and the quantized model using zero-shot quantization. The second comparison is between using the sum of mean\_loss and standard\_deviation\_loss versus using only mean\_loss for backpropagation in data distillation. The third comparison is between post-training quantization, which requires training data, and zero-shot quantization. All these comparisons were conducted for both RGB images trained RetinaNet and thermal images trained RetinaNet to explore zero-shot quantization in the infrared domain.

The MS-COCO pretrained RetinaNet is used as a base for zero-shot quantization experiments. Only a \%4 mAP degradation occurred after zero-shot quantization, while the model size shrunk by four times. Then, the pretrained RetinaNet is fine-tuned with the FLIR ADAS dataset, achieving an mAP of 32.29. Zero-shot quantization resulted in an mAP of 31.18 after quantization. In the comparison of fine-tuned and pre-trained models, zero-shot quantization performed slightly better in the fine-tuned RetinaNet. Numerical results are provided in Table \ref{tab:acc-size-results}, demonstrating the successful application of zero-shot quantization. Since we decreased the output number of the classification head in fine-tuning, the model size slightly decreased after fine-tuning.

We conducted an experiment to evaluate the effect of the number of iterations used for data distillation on quantization accuracy. For each specified number of iterations, we repeated the distillation process 128 times to obtain a batch of 128 distilled data samples. These batches were then used for the calibration of quantization. To assess the impact of the number of iterations on quantization accuracy, we measured the mean Average Precision (mAP) for each batch. The results are presented in a graph where the x-axis represents the number of iterations and the y-axis represents the resulting mAP. This approach allowed us to systematically analyze how varying the number of iterations during data distillation influences quantization performance. The findings, illustrated in the graph, provide insights into the optimal number of iterations needed to achieve the best quantization accuracy. 

\begin{figure}[htbp]
    \centering
    \captionsetup{justification=centering, font={normalsize}, width=\textwidth}
    \begin{subfigure}[b]{0.35\textwidth}
        \centering
        \includegraphics[width=\textwidth]{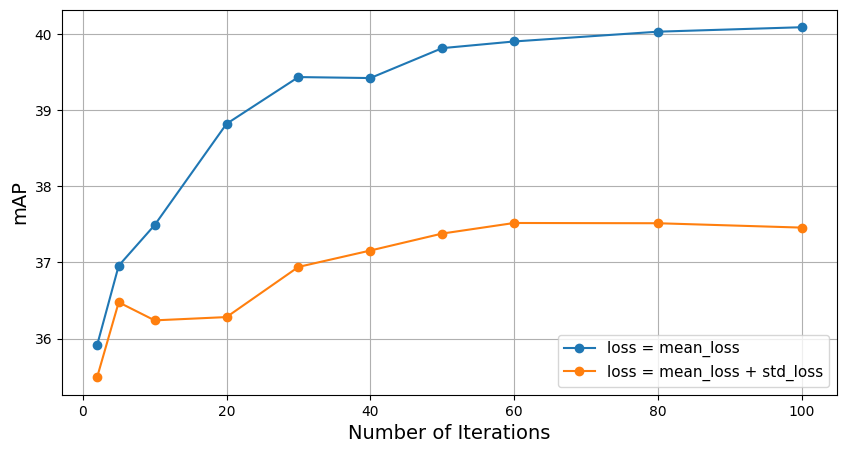}
        \caption{Distilled data accuracy versus the number of iterations during distillation using pretrained RetinaNet.}
        \label{fig:acc_coco}
    \end{subfigure}
    \hspace{0.05\textwidth}
    \begin{subfigure}[b]{0.35\textwidth}
        \centering
        \includegraphics[width=\textwidth]{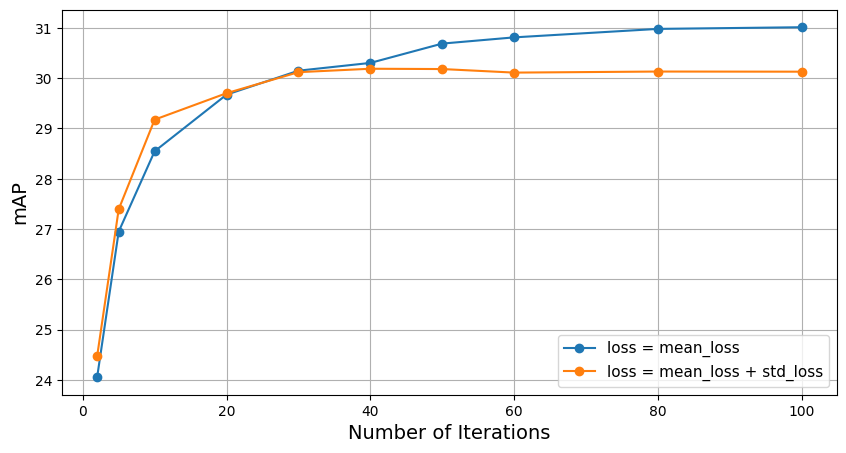}
        \caption{Distilled data accuracy versus the number of iterations during distillation using fine-tuned RetinaNet}
        \label{fig:acc_flir}
    \end{subfigure}
    \vspace{0.3cm} 
    \begin{subfigure}[b]{0.35\textwidth}
        \centering
        \includegraphics[width=\textwidth]{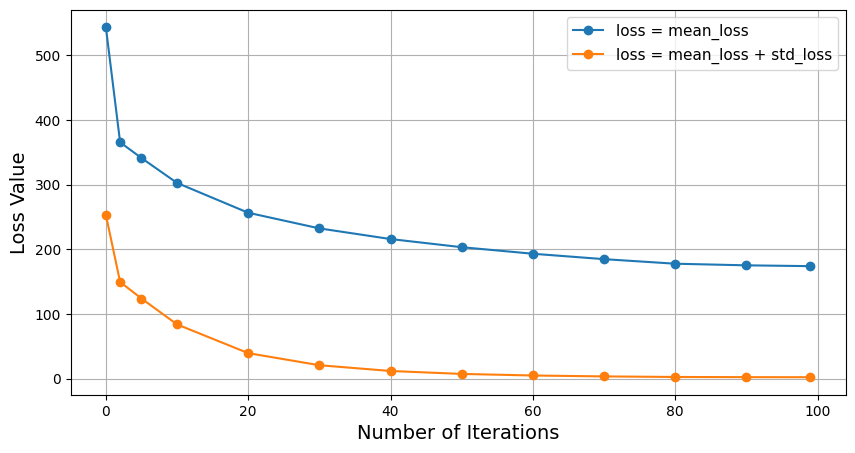}
        \caption{Loss during distillation versus the number of iterations using pretrained RetinaNet}
        \label{fig:loss_coco}
    \end{subfigure}
    \hspace{0.05\textwidth}
    \begin{subfigure}[b]{0.35\textwidth}
        \centering
        \includegraphics[width=\textwidth]{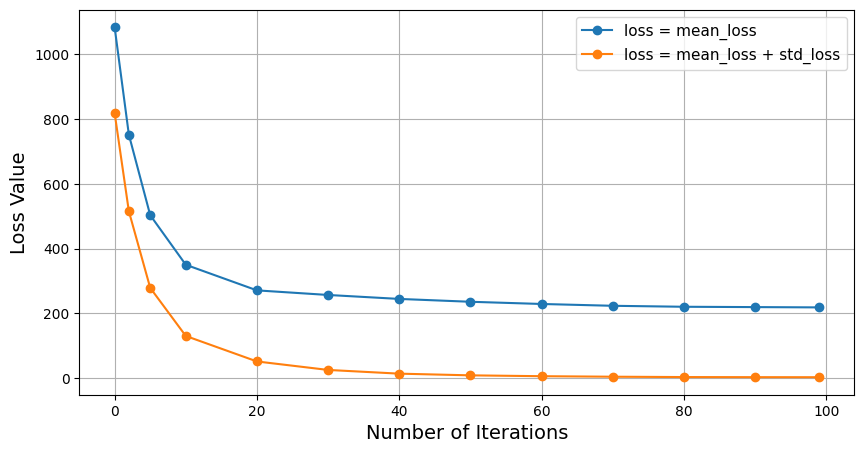}
        \caption{Loss during distillation versus the number of iterations using fine-tuned RetinaNet}
        \label{fig:loss_flir}
    \end{subfigure}
    \vspace{0.1cm} 
    \caption{Comparison of distilled data accuracy and loss during distillation with varying numbers of iterations for both pretrained and fine-tuned RetinaNet models.}
    \label{fig:comparison_acc_loss}
\end{figure}

In the data distillation process, the loss consists of two components: one from the mean statistic of the batch normalization layer (mean\_loss) and the other from the standard deviation statistic (standard\_deviation\_loss). The sum of mean\_loss and standard\_deviation\_loss is typically backpropagated to the distilled data as described in \cite{ZeroQ, hardsamples, intraq}. We initially used this combined loss for backpropagation. However, we observed that using only mean\_loss for backpropagation yields better quantization accuracy. Consequently, our results for comparing full-precision and quantized models, as well as for comparing the training dataset and distilled dataset, are based on using only mean\_loss for backpropagation. The comparison of results from both losses is shown in Figure \ref{fig:comparison_acc_loss}. 
In addition to testing distilled data performance, we evaluated quantization performance using data initialized randomly from a Gaussian distribution with zero mean and unit variance. We observed a mAP of 16.5 for the pretrained RetinaNet and 19.18 for the fine-tuned RetinaNet.

\begin{figure}[htbp]
    \centering
  \captionsetup{justification=centering, font={normalsize}, width=\textwidth}
    \includegraphics[width=0.5\linewidth]{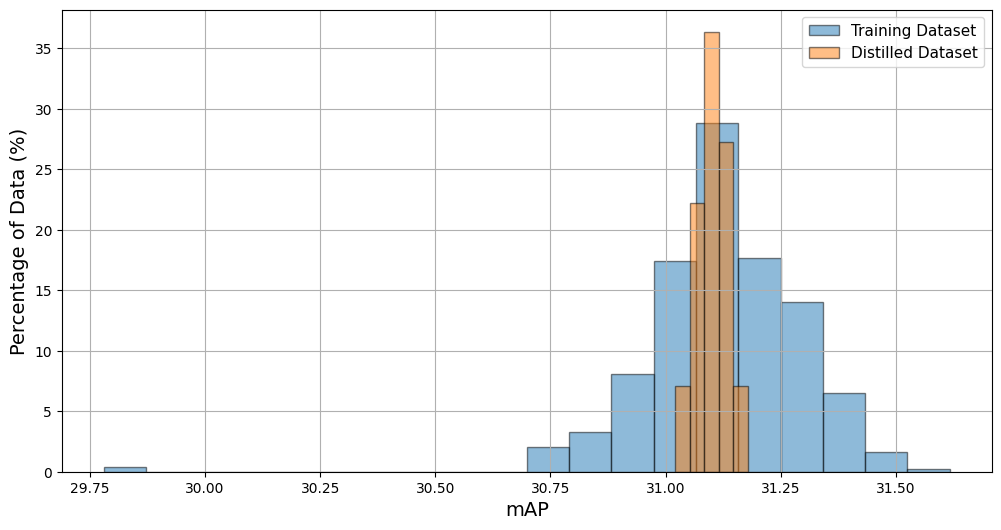}
    \caption{Comparison of training data (post-training quantization) and distilled data (zero-shot quantization) in terms of resulted accuracy.}
    \label{fig:dataset_comp}
\end{figure}

To compare zero-shot quantization performance with post-training quantization, we utilized FLIR ADAS dataset for quantization calibration. Since each data in the dataset does not perform the same, it is important to use a batch of data for calibration. However, data with low performance can also degrade the performance of the batch. Therefore, we created batches with size of 8 from FLIR ADAS training dataset using consecutive images. We then collected statistics of the resulting accuracy from each batch and illustrated them in Figure \ref{fig:dataset_comp}. The mean of the obtained results is 31.13 mAP, with the minimum accuracy recorded at 29.78 mAP. Additionally, we distilled 1024 images and divided them into 8-size batches. We added the corresponding accuracy results to Figure \ref{fig:dataset_comp}. The mean of the obtained results is 31.10 mAP, which is nearly the same as the results of post-training quantization. Also, the deviation in the resulting accuracy of the zero-shot quantization is much smaller than that of the post-training quantization. The  mAP values are rounded for better visualization and comparison in Figure \ref{fig:dataset_comp}.

\begin{figure}[t]
  \centering
  \captionsetup{justification=centering, font={normalsize}, width=\textwidth}
  \begin{subfigure}{0.60\linewidth}
    \includegraphics[width=1.0\linewidth]{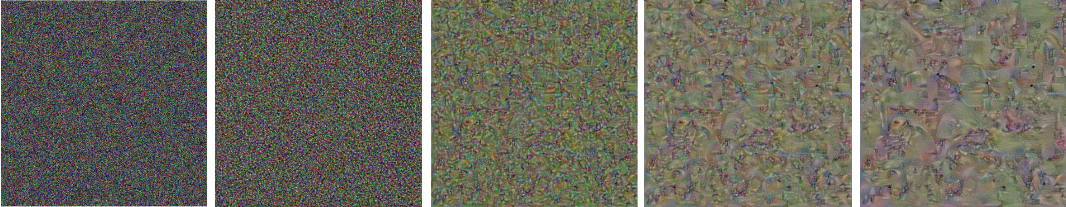}
    \caption{Distilled Data from MS-COCO pretrained RetinaNet}
    \label{fig:rgb-distill}
  \end{subfigure}
  \hfill
  \begin{subfigure}{0.60\linewidth}
    \includegraphics[width=1.0\linewidth]{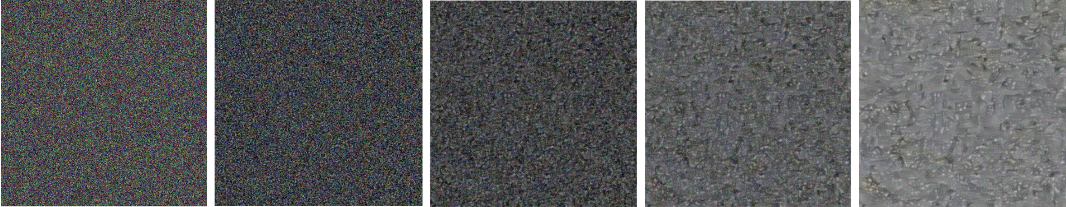}
    \caption{Distilled Data from FLIR ADAS fine-tuned RetinaNet}
    \label{fig:flir-distill}
  \end{subfigure}
  \caption{Accuracy comparison of full precision (32-bit weights, 32-bit activations) and quantized (8-bit weights, 8-bit activations). RetinaNet with pretrained weights and weights after fine-tuning. Zero-shot quantization is applied.}
  \label{fig:distilled-img}
\end{figure}

To illustrate the changes in the distilled data after fine-tuning with thermal images, we visualized the distilled data from both the pretrained RetinaNet and the fine-tuned RetinaNet in Figure \ref{fig:distilled-img}. Distilled data from the fine-tuned RetinaNet is close to gray-scale, since it should represent the statistics of the training dataset. For a deeper understanding of the distilled data, we presented it at various iteration numbers. 


\section{Conclusion}
\label{sec:conclusion}

We have demonstrated that zero-shot quantization using batch-normalization statistics can be successfully applied to fine-tuned models with thermal images. We compared RetinaNet trained on RGB and thermal datasets across various aspects of zero-shot quantization. Our analysis of the number of iterations for data distillation revealed that using only the mean statistic of batch normalization layers for loss backpropagation improved quantization accuracy compared to using both the mean and standard deviation statistics. Additionally, our comparison of zero-shot quantization with post-training quantization illustrated that the mean mAP of zero-shot quantization was nearly identical to that of post-training quantization, with a smaller deviation in accuracy. This consistency underscores the reliability of zero-shot quantization.


\end{document}